\documentclass[10pt,twocolumn,letterpaper]{article}

\usepackage{cvpr}              % To produce the CAMERA-READY version
\definecolor{cvprblue}{rgb}{0.21,0.49,0.74}
\usepackage[pagebackref,breaklinks,colorlinks,allcolors=cvprblue]{hyperref}

\usepackage{algorithm}
\usepackage{algorithmic}
\usepackage{arydshln}
\usepackage{multirow}

\def\paperID{} % *** Enter the Paper ID here
\def\confName{CVPR}
\def\confYear{2026}

\title{Layer Embedding Deep Fusion Graph Neural Network}

\author{
Taihua Xu\textsuperscript{1,*},
Genhao Tian\textsuperscript{1,*}, 
Jicong Fan\textsuperscript{2,$\dagger$},
Xibei Yang\textsuperscript{1},
Qinghua Zhang\textsuperscript{3},
Yun Cui\textsuperscript{1}
\\
{\textsuperscript{1}Jiangsu University of Science and Technology} \quad\\
{\textsuperscript{2}The Chinese University of Hong Kong, Shenzhen} \quad\\
{\textsuperscript{3}Chongqing University of Post and Telecommunications} \\
}

\begin{document}
\maketitle
\let\thefootnote\relax\footnotetext{* Equal contribution.}
\let\thefootnote\relax\footnotetext{$\dagger$ Corresponding author.}
\begin{abstract}
Graph Neural Networks (GNNs) have demonstrated impressive  performance in learning representations from graph-structured data. However, their message-passing mechanism inherently relies on the assumption of label consistency among connected nodes, limiting their applicability to low-homophily settings. 
Moreover, since message passing operates as a hierarchical diffusion process, GNNs face challenges in capturing long-range dependencies. As network depth increases, the structural noise along heterophilic edges tends to be amplified, resulting in over-smoothing.
This issue becomes especially prominent in highly heterophilic graphs, where the propagation of inconsistent semantics across the topology continually exacerbates misaggregation. To address this issue, we propose a novel framework named Layer Embedding Deep Fusion Graph Neural Network (LEDF-GNN).
Specifically, we design a Layer Embedding Deep Fusion (LEDF) operator that nonlinearly fuses multi-layer embeddings to capture inter-layer dependencies and effectively alleviate deep propagation degradation. Meanwhile, to mitigate structural heterophily, LEDF-GNN employs a Dual-Topology Parallel Strategy (DTPS) that simultaneously leverages the original and reconstructed topologies, allowing for adaptive structure–semantics co-optimization under diverse homophily conditions. Extensive semi-supervised classification experiments on the citation and image benchmarks demonstrate that, under both homophilic and heterophilic settings, LEDF-GNN consistently outperforms state-of-the-art baselines, validating its effectiveness and generalization capability across diverse graph types.

\end{abstract}    
\section{Introduction}
As a non-Euclidean structure, graph data is widely present and very powerful in the real world. Several excellent classic GNNs \cite{kipf2017semi,velivckovic2018graph,gasteiger2019predict}  had been developed for graph data. For specific tasks, GNNs can be categorized into node-level \cite{kipf2017semi}, edge-level \cite{zhang2018link,cai2020multi}, and graph-level \cite{han2022g,zhang2018end} methods. All categories of GNNs always need to learn the embedding representations of the graph data.

The core idea of GNNs lies in the message passing mechanism \cite{gilmer2017neural}, which aggregates information within local structural neighborhoods to obtain node embeddings that capture both topological and semantic characteristics. Representative models such as GCN and GAT are built upon this mechanism and have demonstrated outstanding effectiveness on homophilic graphs.

However, the effectiveness of GNNs heavily relies on the implicit label consistency assumption that adjacent nodes share similar labels or semantics \cite{bi2024make,du2022gbk}. In real-world scenarios, graph structures often exhibit heterophily, where connected nodes differ significantly in semantic or label space. Under such conditions, topology-based message passing inevitably introduces semantic noise and label conflicts during aggregation, disrupting the consistency of node representations and degrading model generalization on heterophilic graphs.

In addition, to capture long-range dependencies, some studies attempt to enlarge the receptive field by stacking more message passing layers \cite{li2021deepgcns_pami}.
However, even in homophilic graphs, excessively deep propagation tends to homogenize node representations, leading to the well-known over-smoothing problem \cite{cai2020note,chen2020measuring,zhao2020pairnorm}.
This phenomenon becomes substantially more severe in heterophilic graphs, where the structural connections often contradict semantic similarity. As the network deepens, the noise propagated through heterophilic edges is repeatedly amplified, causing representations to drift away from their intrinsic semantic distribution and resulting in misaggregation.
Consequently, GNNs encounter a dual bottleneck: structural heterophily and deep propagation degradation, together leading to significant performance degradation. The former introduces semantic noise through heterophilic edges, while the latter amplifies this noise during deep propagation.

From a methodological perspective, the root cause of this dual bottleneck lies not in a single structural defect or layer design, but in the inherent limitations of the message passing paradigm itself. It implicitly treats information propagation as a unidirectional hierarchical diffusion process, neglecting the semantic importance and nonlinear dependencies among layers. Therefore, merely refining the topology or deepening the propagation process cannot yield an optimal solution. We argue that overcoming this dual bottleneck requires to establish a structure–semantics co-optimization mechanism, which can simultaneously correct structural noise at the input level and adaptively integrate multi-layer representations at the fusion level to achieve robust information transmission.

To address these challenges, we propose a novel \textit{Layer Embedding Deep Fusion Graph Neural Network} (LEDF-GNN), which enhances both the structure–semantics consistency and representational ability. From the structural perspective, LEDF-GNN introduces a \textit{Dual-Topology Parallel Strategy} (DTPS) that constructs a logical similarity-based reconstructed topology to complement the original graph, and adaptively fuses the two propagation paths at the node level. This strategy mitigates topology-induced misaggregation and improves information reliability in heterophilic or noisy graphs. From the representational perspective, LEDF-GNN employs a deep nonlinear network to fuse embeddings from multiple propagation layers, which implicitly capturing complex inter-layer dependencies beyond convex-weighted or attention-based linear fusion. This design preserves the semantic diversity of shallow layers while enhancing the global expressiveness of deeper embeddings. Consequently, the dual-topology and deep-fusion mechanisms jointly enable LEDF-GNN to achieve robust and generalizable representations on heterophilic and structurally complex graphs.

The contributions are summarized as follows:
\begin{enumerate}

\item We introduce a Dual-Topology Parallel Strategy (DTPS) that employs a Logical Similarity Coefficient (LSC) to reconstruct a semantically aligned topology $A_R$, and adaptively fuses representations from the original and reconstructed topologies via node-wise learned weights, thereby improving effective homophily and reducing misaggregation in heterophilic graphs.

\item We propose a Layer Embedding Deep Fusion (LEDF) operator that stacks multi-layer embeddings into a tensor and applies a deep nonlinear mapping to fuse them. Unlike convex-weighted or attention-based linear fusion, LEDF implicitly models complex inter-layer dependencies, mitigating the over-smoothing problem while preserving local diversity. LEDF is model-agnostic and can be plugged into various GNN backbones.

\item We integrate DTPS and LEDF into a unified structure–semantic co-optimization framework, named LEDF-GNN, which can serve as a plug-in enhancement module for various backbone GNN architectures. Extensive experiments demonstrate that LEDF-GNN consistently improves model robustness and generalization across both homophilic and heterophilic benchmarks.

\end{enumerate}

\section{Related Work}
\textbf{Graph Neural Networks.}
Graph Neural Networks (GNNs) extract high-order semantic features from local neighborhoods through message passing and feature aggregation on graph structures. Typical GNN methods include GCN \cite{kipf2017semi}, GraphSAGE \cite{graphsage}, and GAT \cite{velivckovic2018graph}, which enhance node representations by iteratively aggregating neighbor features. However, the layer-wise aggregation mechanism also introduces two core challenges: first, model performance heavily depends on the label consistency assumption; second, it struggles to effectively capture long-range dependencies. To address these limitations, research has mainly focused on two directions: deep feature fusion and topology optimization.

\textbf{Deep Graph Neural Networks.}
To mitigate issues such as over-smoothing and feature degradation in deep GNNs, existing studies primarily adopt two types of strategies.
The first type is cross-layer embedding fusion: JKNet \cite{xu2018representation} proposed cross-layer fusion strategy that aggregate different layer embeddings to enhance the diversity and stability of the overall feature representation.
The second type is deep structural optimization: GCNII \cite{chen2020simple} leveraged initial residual connections and identity mapping to enable deep architectures while preserving node self-features; NDLS \cite{zhang2021node} assigned different propagation depths to nodes based on differences in their neighborhood structures to reduce uneven information propagation; DropEdge \cite{rong2019dropedge} randomly removed edges to break over-dependence on graph structure, thereby alleviating over-smoothing.

\textbf{Multi Topology Strategy.}
Beyond deep structural optimization, another line of research focuses on reconstructing and enhancing graph topologies. AM-GCN \cite{wang2020gcn} constructs feature-based topologies using $k$NN based on feature similarity, and performs convolutions on both the original and feature topologies to capture richer feature information. DaGCN \cite{huang2024nodes} integrates virtual features generated by a variational autoencoder with original features to construct a “dual-similarity” topology, enhancing the graph representation capacity. Geom-GCN \cite{geomgcn} approaches topology from a geometric perspective, modeling node relationships in latent space with various embedding methods and performing convolutions on the combined feature and structural topologies.

\section{Methodology}
This section consists of four subsections: Analysis, Notation, Layer Embedding Deep Fusion(LEDF) and LEDF-GNN.

\subsection{Analysis}

In this subsection, we discuss the effects of message passing on homophilic and heterophilic graphs.

\textbf{Message Passing Mechanism and Topology–Semantics Alignment.}
In GNNs, message passing mechanism enables limited labeled nodes to propagate supervisory signals to unlabeled ones by aggregating node features through the graph topology, thereby enhancing semi-supervised learning performance. Consequently, the model  performance heavily depends on the alignment between the topological structure and the task semantics. Only when the graph structure is reasonably aligned with the label distribution can message passing facilitate meaningful feature aggregation. Conversely, if the topology and the label space are inconsistent, the propagated information may lead to misaggregation, ultimately degrading classification performance.

\textbf{Intra-Class and Inter-Class Smoothing.}
Graph signal smoothing can be categorized into \textit{intra-class smoothing} and \textit{inter-class smoothing}.  
In the ideal case of intra-class smoothing, features from nodes within the same class gradually converge to a common class center, enhancing intra-class consistency and inter-class separability. Even when “over-smoothing” occurs (i.e., complete convergence within each class), classification performance remains unaffected.  
However, when a large number of inter-class edges exist, the model would mix features from nodes of different classes during message passing, leading to inter-class smoothing. As a result, heterogeneous features are unwelcome pulled closer together, and originally separable clusters become blended. This phenomenon fundamentally explains why over-smoothing is more severe in heterophilic graphs.

\textbf{Over-Smoothing Differences in Homophilic and Heterophilic Graphs.}
In highly homophilic graphs, even with a few inter-class edges, after sufficient layers of message passing, according to the Markov chain convergence theorem, the degree-normalized adjacency matrix with self-loops $\hat{A}$ (when connected and aperiodic) satisfies:
$\mathbf{F} \in \mathbb{R}^{n \times d}$
\begin{equation}
    \hat{A}^{l} = \mathbf{1\Pi}^{\top}, \quad l \to \infty,
\end{equation}
where $\mathbf{\Pi}$ denotes the stationary distribution vector, $\mathbf{1} \in \mathbb{R}^n$ and $\mathbf{1\Pi}^{\top}$ is a Rank-1 matrix. This indicates that all node features eventually converge toward the same direction, forming \textit{global smoothing}.  
In contrast, in highly heterophilic graphs, most edges connect nodes from different classes. As nodes mainly aggregate features from dissimilar neighbors, this \textit{misaggregation} rapidly destroys feature discriminability within only a few propagation rounds, leading to a drastic performance drop.

\begin{figure}[!ht]
    \centering
    \includegraphics[width=\linewidth]{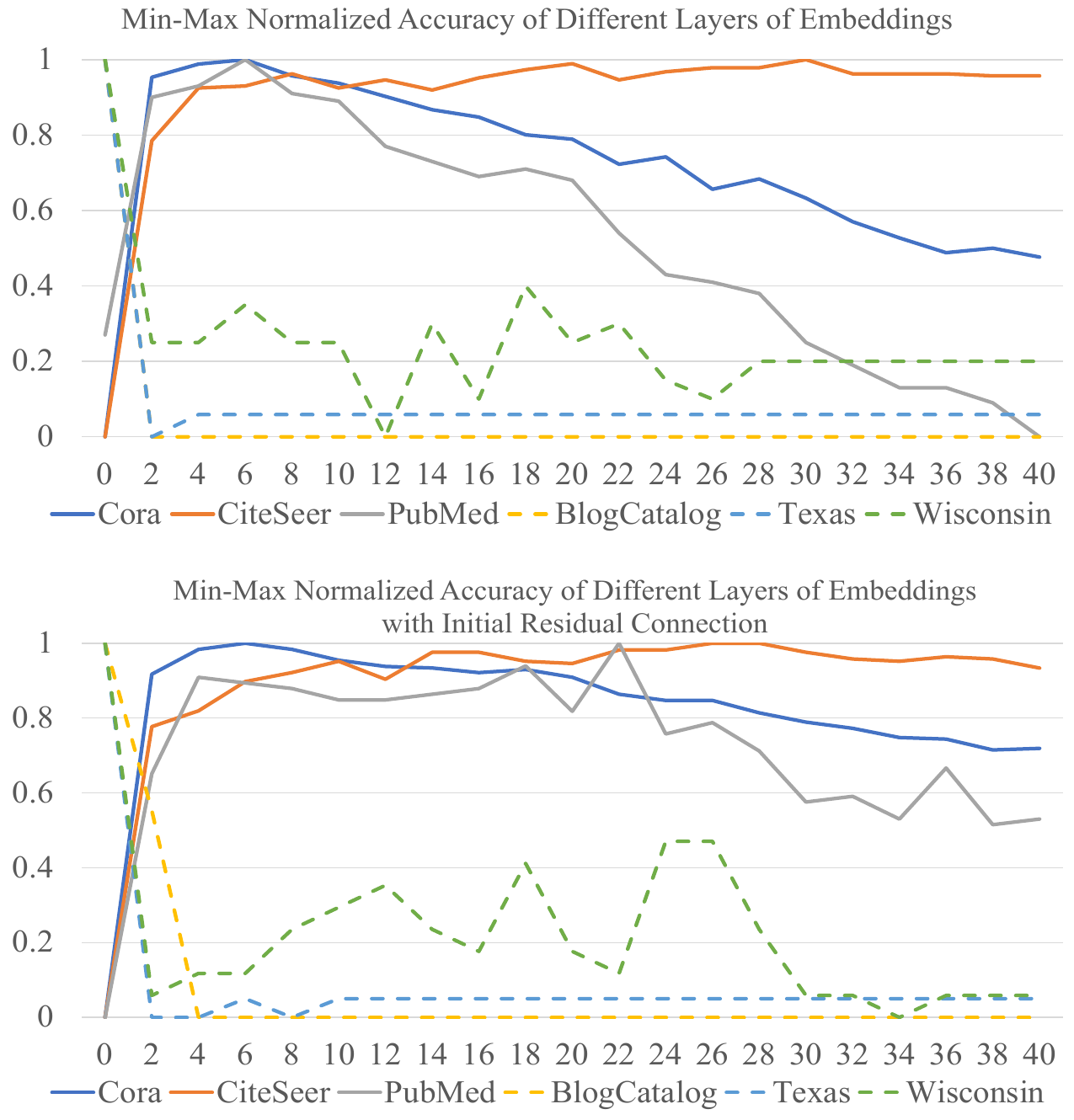}
    \caption{Min-Max normalized classification accuracy of homophilic datasets (Cora, CiteSeer, and PubMed) and heterophilic datasets (BlogCatalog, Texas, and Wisconsin)  across different rounds of message passing. The homophilic datasets are represented by solid lines and the heterophilic ones by dashed line. The x-axis denotes the number of propagation rounds. }
    \label{smooth}
\end{figure}

\textbf{Limitations of Initial Residual Connections and Motivation for the Dual-Topology Strategy.}
To mitigate over-smoothing, \textit{Initial Residual Connections (IRC)} are commonly adopted. We conduct two sets of comparative experiments on both homophilic and heterophilic datasets, using the following formulation:

\begin{equation}
    \mathbf{\hat{Y}} = \mathrm{MLP}(\hat{\mathbf{A}}^{l}\mathbf{F} + \mathrm{diag}(\epsilon) \cdot \mathbf{F}),
\end{equation}
where $\epsilon = 0.1$ and $\hat{\mathbf{A}}$ denotes the degree-normalized adjacency matrix with self-loops.  
The experimental results (see Fig.~\ref{smooth}) show that:
\begin{itemize}
    \item Regardless of the degree of homophily, excessive message passing leads to over-smoothing and subsequent performance degradation;
    \item Over-smoothing occurs earlier and more destructively in heterophilic data (dashed lines);
    \item Initial residual connections fail to significantly alleviate misaggregation in heterophilic settings.
\end{itemize}

These findings suggest that the root cause of over-smoothing in heterophilic graphs lies in \textit{misaggregation} rather than mere feature convergence. Therefore, residual mechanisms alone are insufficient to address this issue effectively.  
Moreover, in practical semi-supervised node classification scenarios, the labels of target nodes are unknown, making it impossible to determine whether local graph regions are homophilic or heterophilic. Consequently, one cannot predict whether message passing will yield beneficial intra-class smoothing or harmful inter-class misaggregation.  
To address this, we propose a \textit{Dual-Topology Parallel Strategy (DTPS)} that jointly leverages both the original and reconstructed topologies. This approach enables the model to adaptively handle varying homophily conditions without requiring prior label information, thereby effectively mitigating misaggregation in heterophilic graphs. We will introduce its implementation in a later section.

\subsection{Notation}
Given a graph $\mathbf{G} = (\mathbf{V},\mathbf{E})$, $\mathbf{V}$ denotes the node set and $\mathbf{E}$ denotes the edge set. 
The original features of all nodes are represented by a matrix $\mathbf{F} \in \mathbb{R}^{n \times d}$ where $n=\left| \mathbf{V}\right|$ is the number of nodes and $d$ is the number of features. Each node has only one label, and the label matrix is $\mathbf{Y} \in \mathbb{R}^{n \times c}$, where \textit{c} is the number of label classes. The prediction of the GNN model is indicated as $\mathbf{P} \in \mathbb{R}^{n \times c}$. The topology structure is represented as an undirected adjacency matrix $\mathbf{A} \in \mathbb{R}^{n \times n}$. The adjacency matrix with self-loop is $\mathbf{\tilde{{A}}} = \mathbf{A}+I$. The degrees of all nodes are represented as a diagonal matrix $\mathbf{D}=diag(d_{1},d_{2},d_{3}, \dots, d_{n})$, where $d_{i}$ is the degree of node $i$. Then the normalized adjacency matrix $\mathbf{\hat{A}}$ is then given by $\mathbf{\hat{A}}=\mathbf{D}^{-\frac{1}{2}}\mathbf{\tilde{A}}\mathbf{D}^{-\frac{1}{2}}$.

\subsection{Layer Embedding Deep Fusion}

In this section, we introduce the \textit{Layer Embedding Deep Fusion (LEDF)} operator, which serves as the core component of our framework. The LEDF operator takes the graph topology and preliminary node representations as inputs, and performs a series of computations to generate fused multi-layer representations. The complete \textit{Layer Embedding Deep Fusion Graph Neural Network (LEDF-GNN)} will be described in the next section.

Before introducing the LEDF operator, we first note that any backbone GNN model can be formulated as:
% \begin{equation}
%     P = GNN(F,\hat{A})
% \end{equation}
\begin{equation}
    \mathbf{P} = \text{GNN}_\theta(\mathbf{F},\hat{\mathbf{A}})
\end{equation}
where $F$ and $\hat{A}$ denote the input features and graph topology, respectively, and $P$ represents the logits produced by the backbone model.

In our model, message passing is performed using the degree-normalized adjacency matrix. The embedding of the $l$-th layer is computed as:
% \begin{equation}
%     P^{(l)} = \hat{A}^{l}P, \quad l \in [1, L],
%     \label{Eq_Fl}
% \end{equation}
\begin{equation}
    \mathbf{P}^{(l)} = \hat{\mathbf{A}}^{q}\mathbf{P}, \quad q \in \{1,2,\ldots, Q\},
    \label{Eq_Fl}
\end{equation}
where $\hat{\mathbf{A}}$ denotes the degree-normalized adjacency matrix with self-loops.
We then stack all layer embeddings into a three-order tensor:

\begin{equation}
    \boldsymbol{\mathcal{X}} = [\mathbf{P}; \mathbf{P}^{(1)}; \dots, \mathbf{P}^{(Q)}]\in \mathbb{R}^{n \times c \times (Q+1)}.
    \label{Eq-X}
\end{equation}

Existing fixed fusion strategies, such as concatenation or max-pooling, treat all layers equally and lack flexibility across datasets. The proposed LEDF operator aims to adaptively fuse the embeddings from multiple layers by employing a nonlinear deep fusion strategy that consists of $K$ fully connected layers, enabling the model to capture both local and global structural information, and allowing the model to learn complex layer-wise dependencies.
Formally, the LEDF operator is defined as:

\begin{equation}
    \mathbf{H} = \sigma(\sigma(\boldsymbol{\mathcal{X}}\times_3\mathbf{W}_{1}) \dots \times_3\mathbf{W}_{K-1})\times_3\mathbf{W}_{K},
    \label{lidf}
\end{equation}
where $\times_j$ denotes the product between a matrix and a matrix along the mode $j$ of the tensor, $\sigma(\cdot)$ denotes the nonlinear activation function, and we adopt ReLU in our experiments. Particularly, $\mathbf{W}_{1}\in\mathbb{R}^{(Q+1)\times \frac{(Q+1)(K-1)}{K}}$ and $\mathbf{W}_{K}\in\mathbb{R}^{\frac{(Q+1)}{K}\times 1}$, which means $\mathbf{H}\in\mathbb{R}^{n\times c}$.

Notably, existing layer fusion methods (such as attention-based weighted summation) typically adopt a convex weighting mechanism to integrate embeddings from different layers, essentially achieving inter-layer information aggregation through a linear combination. 
However, such approaches face limitations in heterophilic graphs: attention weights tend to bias toward more reliable shallow layers, as deep embeddings can be easily corrupted by noise and semantic distortion. This weight collapse leads to an over-reliance on shallow semantics and weakens the modeling of nonlinear inter-layer dependencies. This tendency is empirically validated by the analysis results presented in the appendix.

The LEDF operator can mitigate the weight collapse problem inherent in convex weighting schemes, which arises from that its ability to effectively establish a trainable high-dimensional representation space across layers, enabling the model to automatically capture semantic dependencies and complementary information among embeddings at different depths.

More importantly, the nonlinear inter-layer modeling endowed by LEDF enhances the model's adaptability and expressive capacity in heterophilic graphs, allowing it to maintain robust feature fusion performance even under highly heterogeneous topological conditions. This advantage will be further demonstrated in the subsequent experimental section.

\subsection{LEDF-GNN}

\begin{figure*}[!ht]
\centering
\includegraphics[width=0.9\textwidth]{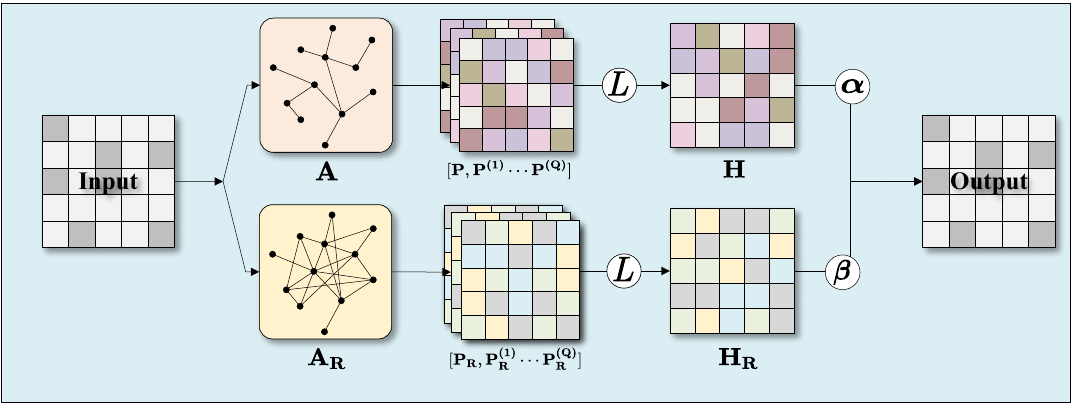}
\caption{ The pipeline of LEDF-GNN. \textcircled{$\mathit{\mathbf{\textit{L}}}$} is the LEDF operator, while 
\textcircled{${\boldsymbol{\alpha}}$} and \textcircled{$\boldsymbol{\beta}$} are the node-wise weight matrices of embedding representations corresponding to the original and reconstructed topologies, respectively. 
}
\label{fig3}
\end{figure*}

Based on the conclusions drawn in the \textit{Analysis} section, the over-smoothing phenomenon observed in heterophilic graphs is primarily caused by \textbf{misaggregation}, rather than mere feature convergence. 
Essentially, misaggregation arises from the existence of numerous \textit{heterophilic edges} in the original low-quality topology, which propagate semantically inconsistent information during message passing. 
As a result, node representations and homophilic information are gradually overwhelmed by a large amount of heterophilic noise. 
Therefore, directly correcting the \textbf{low-quality topology} of heterophilic graphs is a straightforward and highly effective way to mitigate over-smoothing.

However, in practical semi-supervised node classification scenarios, the homophily level of the target dataset is usually unknown. 
Hence, it is impossible to determine whether message passing based on the given topology will produce beneficial intra-class smoothing or harmful inter-class misaggregation. 
To address this challenge, we propose a \textbf{Dual-Topology Parallel Strategy (DTPS)} in LEDF-GNN, which jointly exploits both the original and the reconstructed topologies. 
Through backpropagation, the model automatically learns to assign adaptive weights to the two topological paths, enabling it to generalize across datasets with varying homophily conditions.

To ensure the quality of the reconstructed topology, we further introduce a novel node similarity measure, termed the \textbf{Logical Similarity Coefficient (LSC)}. 
Unlike cosine similarity that focuses on directional consistency or Euclidean distance that measures geometric proximity, LSC emphasizes the \textit{logical commonality and difference} between discrete-valued feature vectors. 
It is defined as:
\begin{equation}
    \mathbf{LSC}_{i,j} = \sum (\mathbf{F}_{i} \wedge \mathbf{F}_{j}) - \gamma \sum (\mathbf{F}_{i} \oplus \mathbf{F}_{j}),
    \label{lsc}
\end{equation}
where $\wedge$ and $\oplus$ denote element-wise logical AND and XOR operations, respectively, and $\gamma$ is a balancing coefficient that penalizes dissimilar components. 
This formulation allows LSC to capture the overlap of active feature dimensions between nodes while suppressing inconsistent dimensions, thereby providing a discrete, interpretable, and noise-robust measure for topology reconstruction.

Since the node features of some graph datasets are continuous values and cannot be directly used for logical operations, it is necessary to first discretize these continuous features. The process is as follows:

\begin{equation}
    f_{i} = 
    \begin{cases}
     0,& abs(f_{i}) - mean(abs(f))\not> 0
     \\
     1,& abs(f_{i}) - mean(abs(f))>0
    \end{cases}
    \label{discrete}
\end{equation}
where $abs(\cdot)$ means taking absolute value.

After having logical similarity between each two nodes, the reconstructed topology is generated from the original topology through logical similarities, and the reconstruction process is defined as follows.
\begin{equation}
    (\mathbf{A_{R}})_{i,j} = \begin{cases}
     \mathbf{A}_{i,j},&LSC_{i,j} \notin Top_k(\mathbf{LSC}_{i})\\
     1,&LSC_{i,j} \in Top_k(\mathbf{LSC}_{i})
    \end{cases}
    \label{AR}
\end{equation}

Based on the reconstructed topology and original topology, we designed a parallel dual-channel structure with one processing the original graph structure and the other processing the reconstructed graph structure. The two channels will return two kinds of embedding representations, $\mathbf{H}$ and $\mathbf{H_R}$, which complement each other, allowing the model to achieve better performance and generalization across different types of data. The final embedding is then obtained by the dual-channel fusion of $\mathbf{H}$ and $\mathbf{H_R}$:
\begin{equation}
    \mathbf{H_{out}} = diag(\alpha) \cdot \mathbf{H} + diag(\beta) \cdot \mathbf{H_{R}}+diag(\epsilon)\cdot \mathbf{P},
\end{equation}
where $diag(\cdot)$ means diagonalization, $\mathbf{P}$ means the initial residual term and $\epsilon = 0.1$; $\alpha$ is the weight parameter of the original topology embeddings, $\beta$ is the weight parameter of the reconstructed topology embeddings, as follows:

\begin{equation}
\boldsymbol{\alpha} =\frac{\exp(\sigma(\mathbf{H}\mathbf{W_{ 1}})\mathbf{W_{ 2}})}{\exp\left( \sigma(\mathbf{H}\mathbf{W_{ 1}})\mathbf{W_{ 2}} 
\right) + \exp\left( \sigma(\mathbf{H_{R}}\mathbf{W_{ 1}})\mathbf{W_{ 2}}  \right)},
\label{t-alpha}
\end{equation}
\begin{equation}
    \boldsymbol{\beta} = \frac{exp(\sigma(\mathbf{H_{R}}\mathbf{W_{ 1}})\mathbf{W_{ 2}})}{\exp\left( \sigma(\mathbf{H}\mathbf{W_{ 1}})\mathbf{W_{ 2}}  \right) + \exp\left( \sigma(\mathbf{H_{R}}\mathbf{W_{1}})\mathbf{W_{2}}  \right)},
\label{t-beta}
\end{equation}
where $\mathbf{W_{1}} \in \mathbb{R}^{c \times \frac{c}{2}}$ and $\mathbf{W_{2}}\in \mathbb{R}^{\frac{c}{2}\times 1}$.

The final prediction $\mathbf{Y}$ of LEDF-GNN is obtained by \begin{equation}
    \mathbf{Y} = SoftMax(\mathbf{H_{out}}).
\end{equation}.

The overall architecture of LEDF-GNN can thus be summarized as a three-stage process:
\begin{enumerate}
    \item \textbf{Representation Propagation:} Generate \textbf{multi-layer} embeddings $\{\mathbf{P}^{(q)}\}$ through degree-normalized message passing.
    \item \textbf{Deep Adaptive Fusion:} Fuse the multi-layer embeddings via the LEDF operator to obtain $\mathbf{H}$, which is subsequently used for downstream prediction.
    \item \textbf{Dual-Topology Representation Fusion:} Fuse the two embeddings from original topology and reconstructed topology via attention mechanism.
\end{enumerate}

Finally, the overall procedure of the proposed LEDF-GNN is summarized in Algorithm \ref{algledf}. 
\begin{algorithm}[h]
\caption{LEDF-GNN}
\label{algledf}
\textbf{Input:} Original features matrix $\mathbf{F}$, normalized adjacency matrix $\hat{\mathbf{A}}$, normalized revised adjacency matrix $\hat{\mathbf{A}}_{R}$, basic GNN model $GNN$.\\
\textbf{Parameter:} $Q1$ and $Q2$. \\
\text{\textbf{Output:} The final prediction $\mathbf{Y}$.}
\begin{algorithmic}[1]
\STATE $\mathbf{P} = GNN(\mathbf{F},\hat{\mathbf{A}})$
\STATE $\mathbf{P}^{(q)} = \hat{\mathbf{A}}^{q}\mathbf{P}, q\in [1,Q_1]$
\STATE $\mathbf{P_{R}}^{(q)} = \hat{\mathbf{A}}^{q}_{R}\mathbf{P}, q\in [1,Q_2]$

\STATE $\boldsymbol{\mathcal{X}} = [\mathbf{P}; \mathbf{P}^{(1)}; \dots, \mathbf{P}^{(Q_1)}]\in \mathbb{R}^{n \times c \times (Q_1+1)}$
% \STATE $X_{R} = [P,P^{(1)},...,P^{(L2)}]_{dim=3}$
\STATE $\boldsymbol{\mathcal{X_R}} = [\mathbf{P_R}; \mathbf{P_R}^{(1)}; \dots, \mathbf{P_R}^{(Q_2)}]\in \mathbb{R}^{n \times c \times (Q_2+1)}$

\STATE $\mathbf{H}= \sigma(\sigma(\boldsymbol{\mathcal{X}}\times_3\mathbf{W}_{1}) \dots \times_3\mathbf{W}_{K-1})\times_2\mathbf{W}_{K}$
 %\STATE $H_{R} = \sigma(\sigma(X_{R}W_{1})...W_{K-1})W_{K}$
\STATE $\mathbf{H_R} = \sigma(\sigma(\boldsymbol{\mathcal{X_R}}\times_3\mathbf{W}^{'}_{1}) \dots \times_3\mathbf{W}^{'}_{K-1})\times_2\mathbf{W}^{'}_{K}$
% \STATE $Y=diag(\boldsymbol{\alpha}) \cdot H + diag(\boldsymbol{\beta})\cdot H_{R}+diag(\epsilon) \cdot P$
\STATE $\mathbf{H_{out}} = diag(\alpha) \cdot \mathbf{H} + diag(\beta) \cdot \mathbf{H_{R}}+diag(\epsilon)\cdot \mathbf{P}$
\STATE $\mathbf{Y} = SoftMax(\mathbf{H_{out}})$
\STATE \textbf{return} $Y$
\end{algorithmic}
\end{algorithm}

\section{Experiment}
This section is consisted of Setup and Baselines, Node Classification Experiment, Ablation Experiment, Validation Experiment and Scalability Analysis subsections.

\subsection{Setup and Baselines}

In the proposed LEDF-GNN, we optimize all parameters using Adam optimization \cite{kingma2014adam} and the cross-entropy loss function. The learning rate is set to 0.01, the weight decay to 0.0005, the dropout to 0.5, and the number of units in hidden backbone to 128.

Because the backbone in LEDF-GNN can be any chosen GNN models, some latest and SOTA plug-in methods are chosen as baselines: BORF \cite{BORF} introduced a Batch Ollivier-Ricci Flow-based rewiring algorithm that simultaneously mitigates both over-smoothing and over-squashing issues, thereby enhancing the expressive capability of GNNs. ComFy \cite{comfy} proposed a hybrid rewiring strategy that leverages both community structure and feature similarity, aiming to strengthen local feature coherence while preserving the global community topology to improve GNN performance. AGMixup \cite{lu2025agmixup} presented an adaptive graph mixup framework that employs data mixing techniques to augment graph representations and further boost the generalization ability of GNNs. The backbones utilized in all methods are as follows: MLP \cite{MLP, MLP2}, GCN \cite{kipf2017semi}, GAT \cite{velivckovic2018graph}, GIN \cite{gin} and APPNP \cite{gasteiger2019predict}.

\subsection{Node Classification Experiment}

In this classification experiment, thirteen datasets are divided into two groups, seven homophilic datasets: Cora, CiteSeer, PubMed\cite{kipf2017semi}, ACM \cite{ACM}, Coauthor-CS \cite{coauthor}, Arxiv2023 \cite{arxiv2023} and MNIST \cite{MNIST}; six heterophilic datasets: BlogCatalog \cite{BlogCatalogFlickr}, Texas, Wisconsin, Cornell \cite{TWC}, Squirrel and Chameleon \cite{Squirrel}.

Table \ref{table4} listed the semi-supervised classification accuracy of all methods across all datasets. The results demonstrate that, compared to other baselines, our model consistently outperform in terms of classification accuracy, both on homophilic and heterophilic datasets.

\begin{table*}[!ht]
\scriptsize 
\centering
\setlength{\tabcolsep}{5pt} 
\begin{tabular}{cccccccc|cccccc}
\hline
Methods&Cora&CiteSeer&PubMed&ACM&CS&Arxiv2023&MNIST&BlogCatalog&Chameleon&Texas&Wisconsin&Cornell&Squrriel \\
\hline
MLP 
&58.5&\underline{60.2}&72.2&67.5&\underline{85.2}&61.9&\underline{75.2}&64.0&32.1&\underline{76.0}&\underline{66.0}&\underline{76.0}&26.6 \\	

$+$ LEDF(ours)
&\textbf{82.5}&\textbf{72.0}&\textbf{80.9}&\textbf{85.5}&\textbf{90.8}&\textbf{77.7}&\textbf{85.5}&\textbf{79.0}&\textbf{43.3}&\textbf{79}&\textbf{69.0}&\textbf{79.0}&\textbf{31.5} \\

$+$ BORF(2023)
&56.2&53.8&70.3&68.1&84.8&-&67.8&57.2&31.5&{72.0}&63.0&72.0&- \\

$+$ ComFy(2025)
&\underline{60.0}&60.0&\underline{72.3}&64.8&84.8&\underline{66.4}&73.6&\underline{66.0}&\underline{36.3}&72.0&63.0&65.0&\underline{26.8} \\
$+$ AGMixup(2025)
&58.0&54.4&72.0&\underline{71.5}&83.4&57.4&62.7&-&35.6&70.0&64.0&66.0&- \\
\hdashline
GCN 
&82.4&70.7&\underline{79.2}&82.1&\underline{89.8}&62.5&82.5&42.0&49.6&55.0&48.0&49.0&\underline{33.7} \\
$+$ LEDF(ours)
&\textbf{84.0}&\textbf{74.8}&\textbf{82.1}&\textbf{87.5}&\textbf{90.5}&\textbf{76.0}&\textbf{86.6}&\textbf{79.6}&\textbf{53.2}&\textbf{66.0}&\textbf{60.0}&\textbf{55.0}&\textbf{39.0} \\
$+$ BORF(2023)
&81.7&69.1&76.7&79.8&89.0&-&82.3&\underline{78.4}&47.6&52.0&50.0&44.0&- \\
$+$ ComFy(2025)
&80.5&\underline{72.3}&79.1&83.2&89.2&\underline{64.0}&82.8&44.8&46.9&\underline{62.0}&\underline{57.0}&\underline{53.0}&\underline{33.7} \\
$+$ AGMixup(2025)
&\underline{82.7}&71.4&78.2&\underline{84.9}&88.4&54.6&\underline{86.5}&-&\underline{51.0}&57.0&48.0&50.0&- \\
\hdashline
GAT
&79.9&70.4&78.0&82.5&87.1&67.3&82.3&43.4&45.7&53.0&47.0&\underline{55.0}&\underline{28.2} \\
$+$ LEDF(ours)
&\textbf{83.7}&\textbf{72.6}&\textbf{81.4}&\textbf{86.4}&\textbf{88.2}&\textbf{81.3}&\textbf{87.1}&\textbf{88.4}&\textbf{52.3}&\textbf{57.0}&\textbf{58.0}&\textbf{56.0}&\textbf{38.0} \\
$+$ BORF(2023)
&80.3&71.2&76.6&81.5&83.6&-&79.4&\underline{52.0}&49.7&43.0&47.0&48.0&- \\
$+$ ComFy(2025)
&\underline{80.8}&70.8&77.8&81.3&85.5&\underline{68.3}&82.2&43.1&45.1&\underline{56.0}&\underline{54.0}&53.0&27.8 \\
$+$ AGMixup(2025)
&79.6&\underline{71.6}&\underline{78.3}&\underline{84.1}&\underline{87.3}&61.1&\underline{84.6}&-&\underline{51.2}&\underline{56.0}&49.0&48.0&- \\
\hdashline
GIN 
&77.1&\underline{67.8}&77.6&\underline{84.5}&83.5&41.6&\underline{83.9}&44.2&49.5&55.0&53.0&\underline{55.0}&\underline{26.9} \\
$+$ LEDF(ours)
&\textbf{81.1}&\textbf{70.9}&\textbf{79.7}&\textbf{87.9}&\textbf{85.7}&\textbf{67.0}&\textbf{88.0}&\textbf{85.9}&\textbf{55.7}&\textbf{65.0}&\textbf{66.0}&\textbf{56.0}&\textbf{35.3} \\
$+$ BORF(2023)
&\underline{77.3}&64.3&\underline{78.1}&80.4&\underline{84.6}&-&81.9&\underline{79.2}&\underline{53.2}&44.0&51.0&44.0&- \\
$+$ ComFy(2025)
&75.7&63.2&77.0&80.7&83.5&\underline{61.4}&82.1&72.8&52.8&50.0&\underline{59.0}&53.0&23.0 \\
$+$ AGMixup(2025)
&76.1&66.6&75.0&78.9&75.4&52.4&74.3&-&52.6&\underline{61.0}&51.0&\underline{55.0}&- \\
\hdashline
APPNP 
&81.5&\underline{71.8}&\underline{79.5}&83.1&90.3&57.0&83.0&58.7&45.3&55.0&48.0&49.0&\underline{31.0} \\
$+$ LEDF(ours)
&\textbf{84.7}&\textbf{74.8}&\textbf{82.4}&\textbf{87.4}&\textbf{91.8}&\textbf{64.4}&\textbf{86.1}&\textbf{91.7}&\textbf{51.2}&\textbf{69.0}&\textbf{57.0}&\textbf{58.0}&\textbf{36.4} \\
$+$ BORF(2023)
& 81.7&69.9&77.0&80.5&90.3&-&76.2&\underline{91.1}&\underline{48.4}&57.0&\underline{52.0}&\underline{56.0}&-  \\
$+$ ComFy(2025)
&79.3&70.4&79.1&82.1&\underline{90.7}&\underline{63.6}&83.4&60.2&43.0&\underline{61.0}&\underline{52.0}&55.0&30.2  \\
$+$ AGMixup(2025)
& \underline{82.5}&71.6&79.0&\underline{83.8}&89.4&58.4&\underline{85.7}&-&47.0&60.0&49.0&51.0&-  \\

\hline
\end{tabular}
\caption{The semi-supervised classification accuracy (\%) of LEDF-GNN all baselines with five GNN backbones across thirteen datasets. \textbf{Bold} is the best, \underline{underline} is the second-best.}
\label{table4}
\end{table*}

\subsection{Ablation Experiment}
To evaluate the necessity and effectiveness of each component in our model, we design four ablation experiments focusing on both DTPS strategy and LEDF operator. The ablation settings are as follows:

\begin{itemize}
    \item \textbf{Ablation 1}: Remove the reconstructed topology and retain only the original topology;
    \item \textbf{Ablation 2}: Remove the original topology and retain only the reconstructed topology;
    \item \textbf{Ablation 3}: Remove the deep multi-layer embedding fusion and retain only the last-layer embedding;
    \item \textbf{Ablation 4}: Remove the deep multi-layer embedding fusion and retain only a middle-layer embedding.
\end{itemize}

\begin{figure}[!h]
    \centering
    \includegraphics[width=\linewidth]{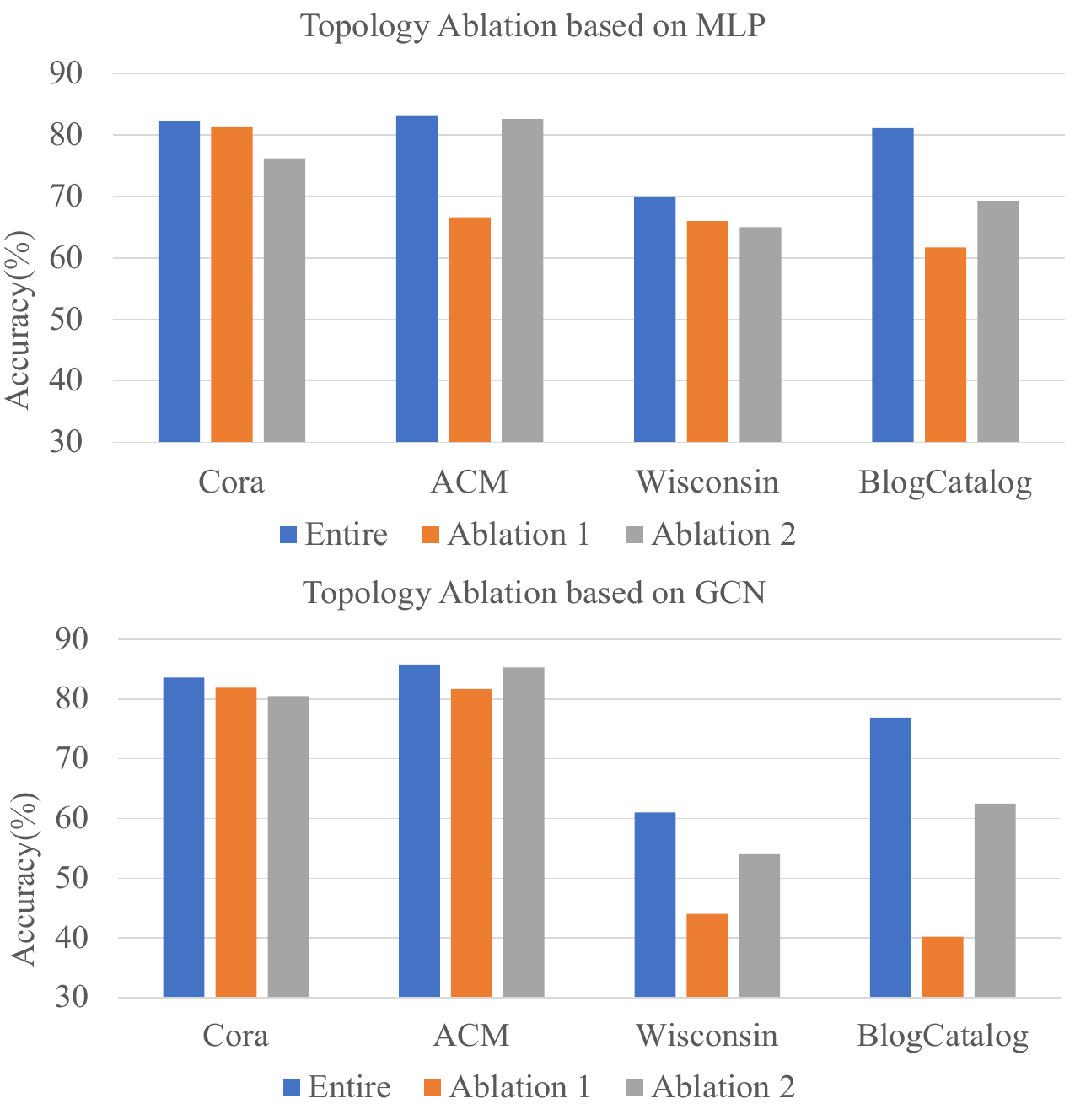}
    \caption{The accuracy contrast of LEDF-GNN and its DTPS ablation version in homophilic datasets (Cora, ACM) and heterophilic datasets (Wisconsin, BlogCatalog) four datasets.}
    \label{ablation topology}
\end{figure}

\begin{figure}[!h]
    \centering
    \includegraphics[width=\linewidth]{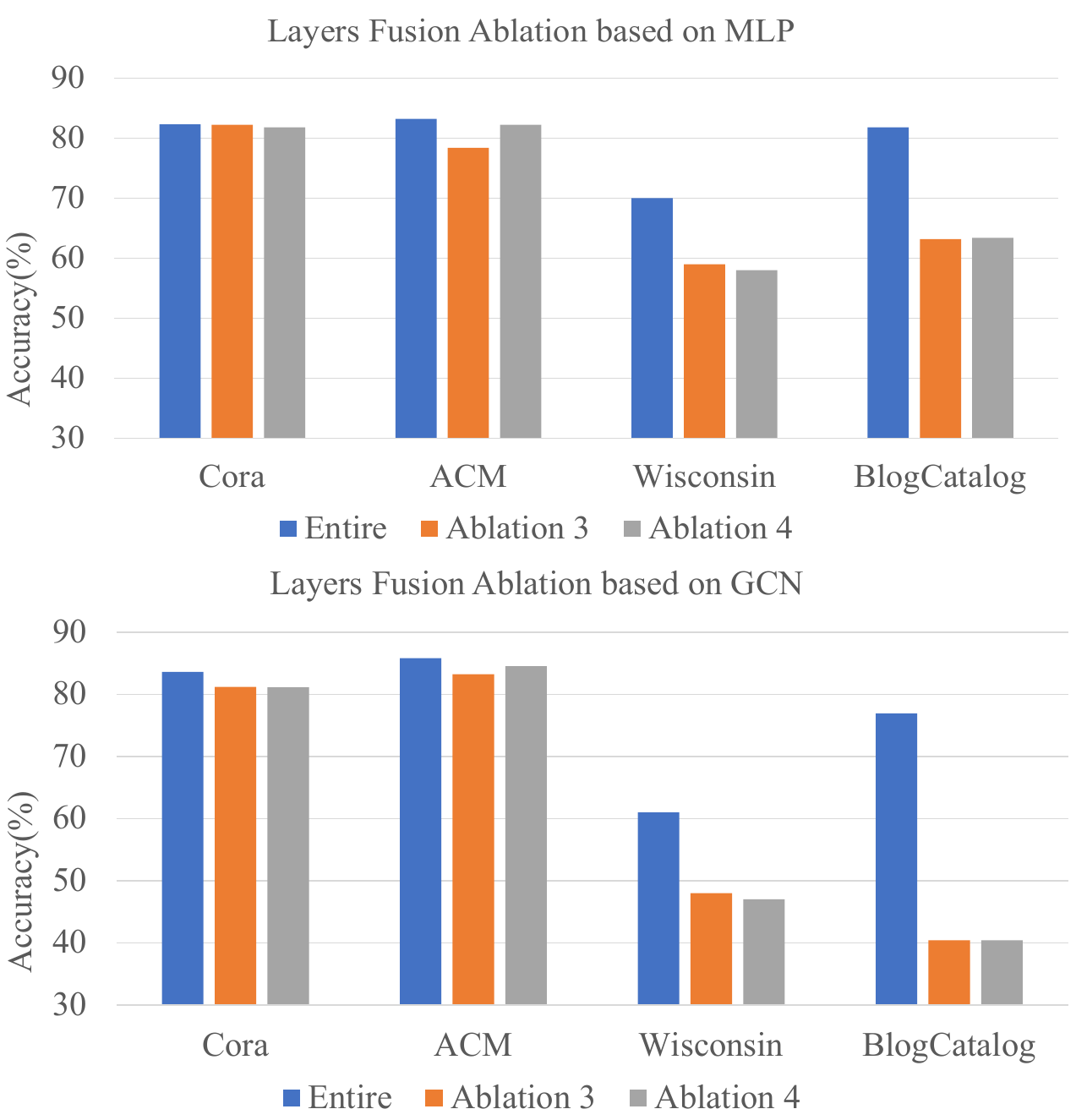}
    \caption{The accuracy contrast of LEDF-GNN and its LEDF ablation version in homophilic datasets (Cora, ACM) and heterophilic datasets (Wisconsin, BlogCatalog) four datasets.}
    \label{ablation fusion}
\end{figure}

Among these, Ablation 1 and Ablation 2 constitute the first group of experiments (shown in Fig. \ref{ablation topology}), aiming to verify the necessity of DTPS strategy. Ablation 3 and Ablation 4 form the second group of experiments (shown in Fig. \ref{ablation fusion}), aiming to validate the effectiveness of the LEDF operator. All experiments are conducted on both homophilic datasets (Cora, ACM) and heterophilic datasets (Wisconsin, BlogCatalog).


Fig. \ref{ablation topology} shows that removing either the original topology or the reconstructed topology leads to noticeable performance degradation, demonstrating the necessity and complementarity of DTPS strategy. Notably, on heterophilic datasets, removing the reconstructed topology while retaining only the original topology (i.e., Ablation 1) results in a significant performance drop. This further supports our earlier argument: in heterophilic scenarios, performance degradation is primarily caused by misaggregation rather than feature homogenization, and the reconstructed topology can effectively alleviate this issue.

Fig. \ref{ablation fusion} shows that retaining only the last-layer or middle-layer embedding consistently leads to decreased performance, confirming the necessity and effectiveness of LEDF operator. 
Moreover, on homophilic datasets, the ablation of LEDF causes a slight performance drop. In contrast, on heterophilic datasets, the ablation of LEDF results in a substantial performance decline  , even though residual technique remains. This observation aligns with our earlier analysis: over-smoothing in heterophilic graphs primarily arises from misaggregation, and initial residual connections alone are unable to counteract such erroneous feature aggregation.

\subsection{Validation Experiments}

\begin{figure}[!h]
    \centering
    \includegraphics[width=\linewidth]{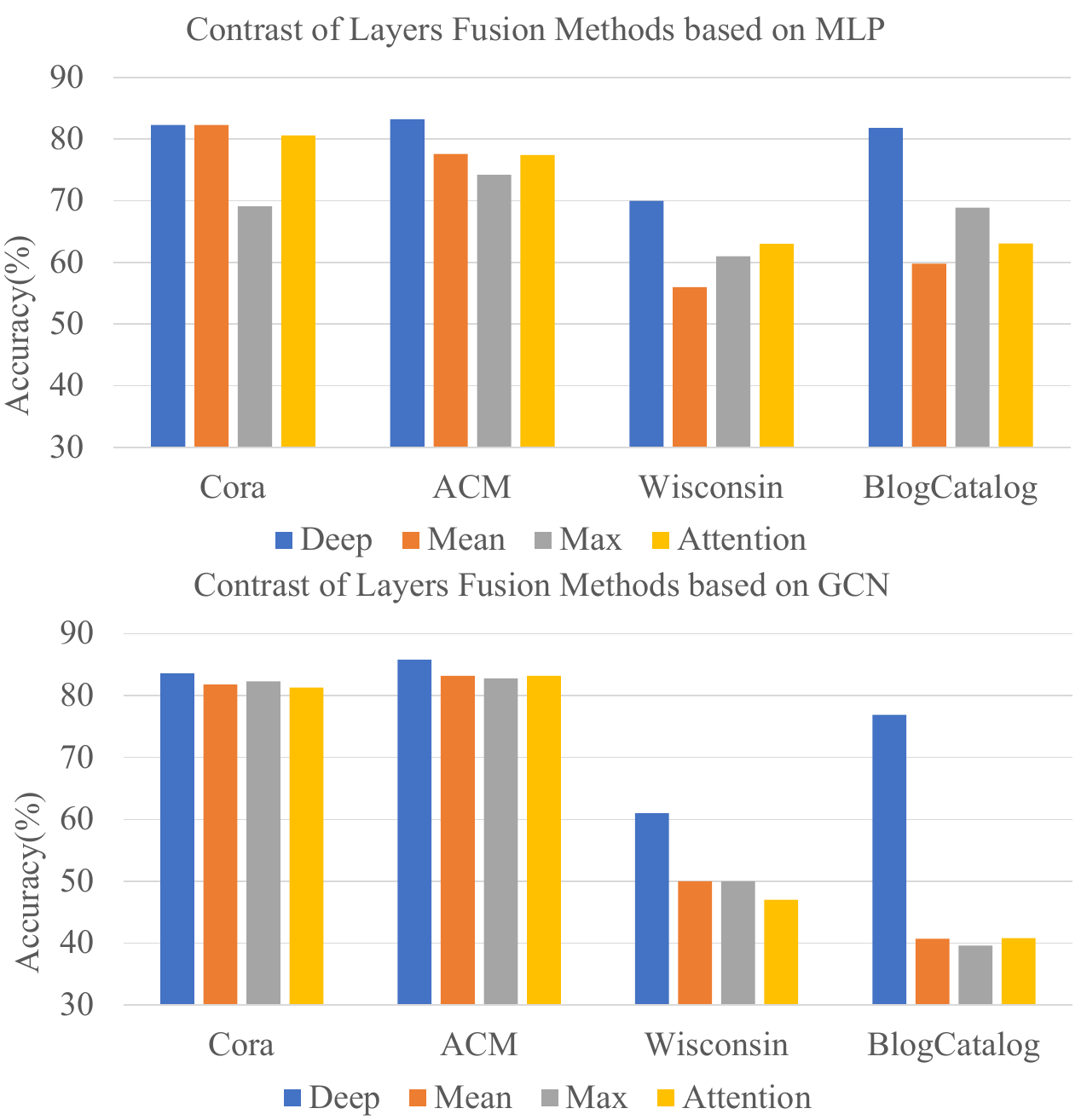}
    \caption{The accuracy contrast of LEDF-GNN and its validation version in homophilic datasets (Cora, ACM) and heterophilic datasets (Wisconsin, BlogCatalog) four datasets.}
    \label{fusion contrast}
\end{figure}

To further evaluate the effectiveness of the proposed LEDF operator, we compare it with three widely used layer fusion strategies: Mean Pooling, Max Pooling, and Attention-based Sum. Experiments are conducted on both homophilic datasets (Cora, ACM) and heterophilic datasets (Wisconsin, BlogCatalog). As shown in Fig. \ref{fusion contrast}, replacing LEDF with other fusion methods leads to a significant performance drop across all datasets, with particularly severity on heterophilic graphs. These results demonstrate that LEDF achieves superior embedding aggregation and generalization capabilities across different types of graph structures.

A deeper analysis reveals that pooling-based methods struggle in heterophilic settings due to high structural noise. The heterophilic connections introduce significant feature variance across layers, making it difficult for statistical pooling operations to distinguish informative signals from irrelevant noise, weakening the overall fusion quality.

Consistent with our earlier analysis, attention-based fusion exhibits a strong bias toward shallow layers in heterophilic graphs. As deep embeddings are easily corrupted by structural noise, the weight distribution collapses toward shallow layers, causing the fused representation to over-rely on shallow semantics and weakening cross-layer integration.

Interestingly, attention mechanisms combined with MLP predictors outperform their GCN-based counterparts in heterophilic settings. This phenomenon can be attributed to the fact that MLP predictors do not rely on message passing and are therefore immune to misaggregation. In contrast, GCN predictors inherently involve message propagation, which amplifies structural noise when convex weighting collapses occur in either shallow or deep layers, leading to severe performance degradation.
These findings further validate our central claim that, in heterophilic graphs, performance degradation primarily arises from erroneous aggregation rather than feature homogenization.

\subsection{Scalability Analysis}
In order to evaluate the scalability of the model, we conducted experiments on the large-scale dataset ogbn-arxiv and achieved certain improvements based on five GNN models. The results are summarized in Table \ref{table_scale}.

\begin{table}[!h]
\centering
\begin{tabular}{ccc}
\hline
&Backbone&Backbone $+$\\
\hline
MLP&38.4&\textbf{66.6}\\
GCN&55.5&\textbf{69.5}\\
GAT&55.1&\textbf{63.5}\\
GIN&52.2&\textbf{64}\\
APPNP&57.4&\textbf{66.8}\\
\hline
\end{tabular}
\caption{The accuracy (\%) of basic GNN backbone and LEDF-GNN on ogbn-arxiv. }
\label{table_scale}
\end{table}

\vspace{-2pt}
\section{Conclusion}

Graph Neural Networks (GNNs) face a dual bottleneck in representation learning on graphs: structural heterophily and deep propagation degradation.
To overcome these issues, we propose LEDF-GNN, a unified framework that jointly enhances depth modeling and structural alignment. The Layer Embedding Deep Fusion (LEDF) operator employs a nonlinear fusion network to adaptively integrate multi-layer embeddings, effectively mitigating propagation degradation. Meanwhile, the Dual-Topology Parallel Strategy (DTPS) constructs a semantically aligned auxiliary topology, and adaptively fuses it with the original graph, alleviating feature misaggregation across heterophilic edges.
Comprehensive experiments on thirteen benchmarks, including ciation and image datasets, demonstrate that LEDF-GNN consistently achieves SOTA performance across both homophilic and heterophilic settings. Ablation studies further confirm the complementary contributions of LEDF and DTPS to address the intrinsic bottlenecks of GNNs.

\section*{Acknowledgements}
This work was supported by the National Natural Science Foundation of China under Grant No.62376236.

{
 \small
 \bibliographystyle{ieeenat_fullname}
 \bibliography{main}
}

\clearpage
\setcounter{page}{1}
\maketitlesupplementary

\section{Attention-based Fusion Method}

To further validate our claim that attention-based layer fusion inherently suffers from weight collapse, especially on heterophilic graphs, we conduct a detailed empirical analysis of attention weights (averaging all nodes) distributions across different propagation depths.

\subsection{Experimental Setup}

We evaluate attention-based multi-layer fusion on four representative datasets, using two backbone predictors: MLP (no message passing) and GCN (involving message passing). To demonstrate the advantage of the reconstructed topology over the original topology in weight collapse, the experimental results on the two topologies are compared side by side for each predictor. 

In particular, we select two homophilic datasets (Cora and ACM) and two heterophilic datasets (Wisconsin and BlogCatalog), ensuring that the evaluation covers both structural regimes. This enables us to comprehensively examine attention weights distribution under different homophily settings.

\subsection{Experiment Result and Analysis}

\begin{figure}[!h]
    \centering
    \includegraphics[width=\linewidth]{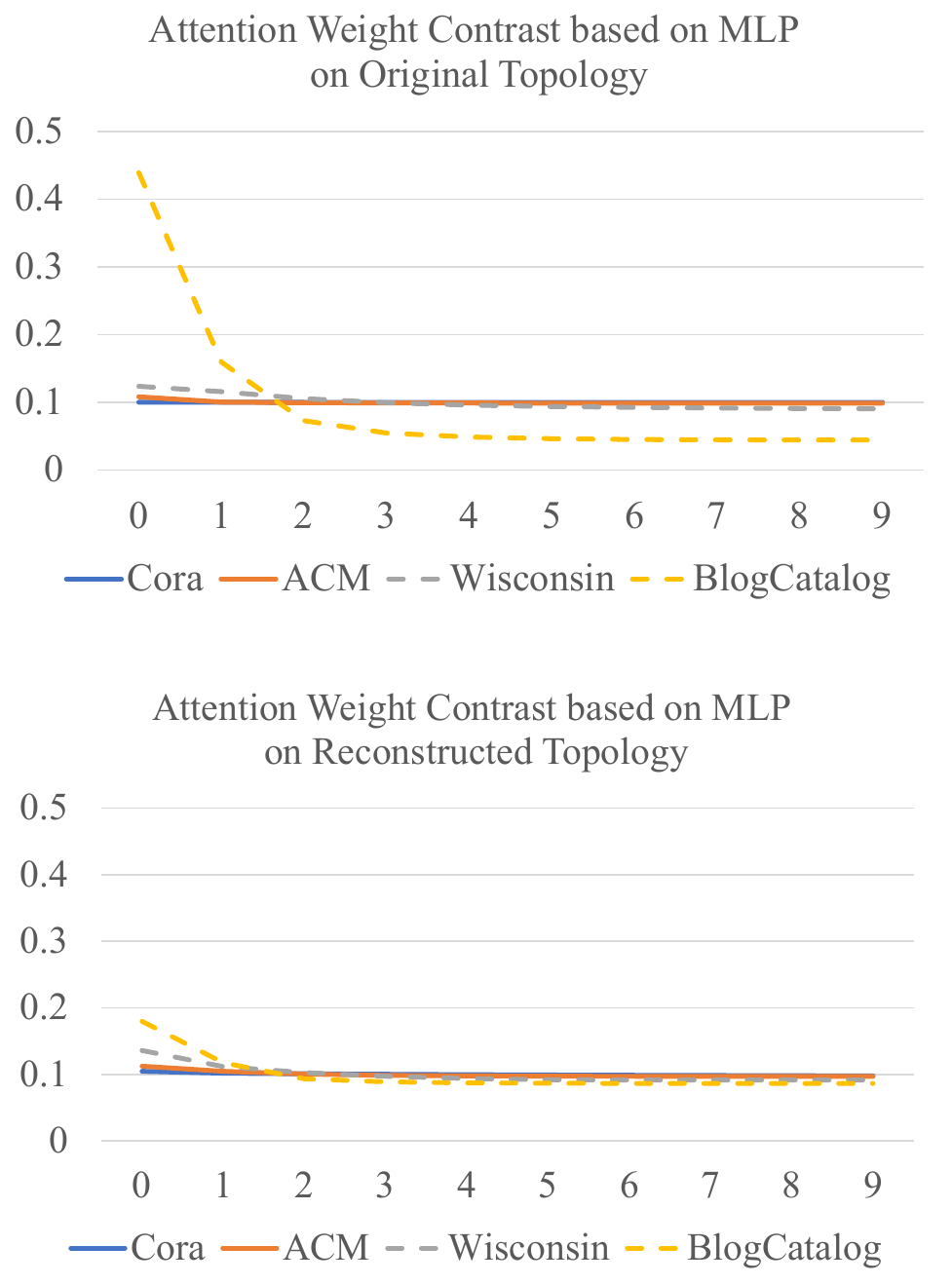}
    \caption{The distribution of attention weights across the layers of the MLP predictor for two different topologies. y-axis is attention wights, x-axis is the massage passing rounds.}
    \label{att1}
\end{figure}

\begin{figure}[!h]
    \centering
    \includegraphics[width=0.9\linewidth]{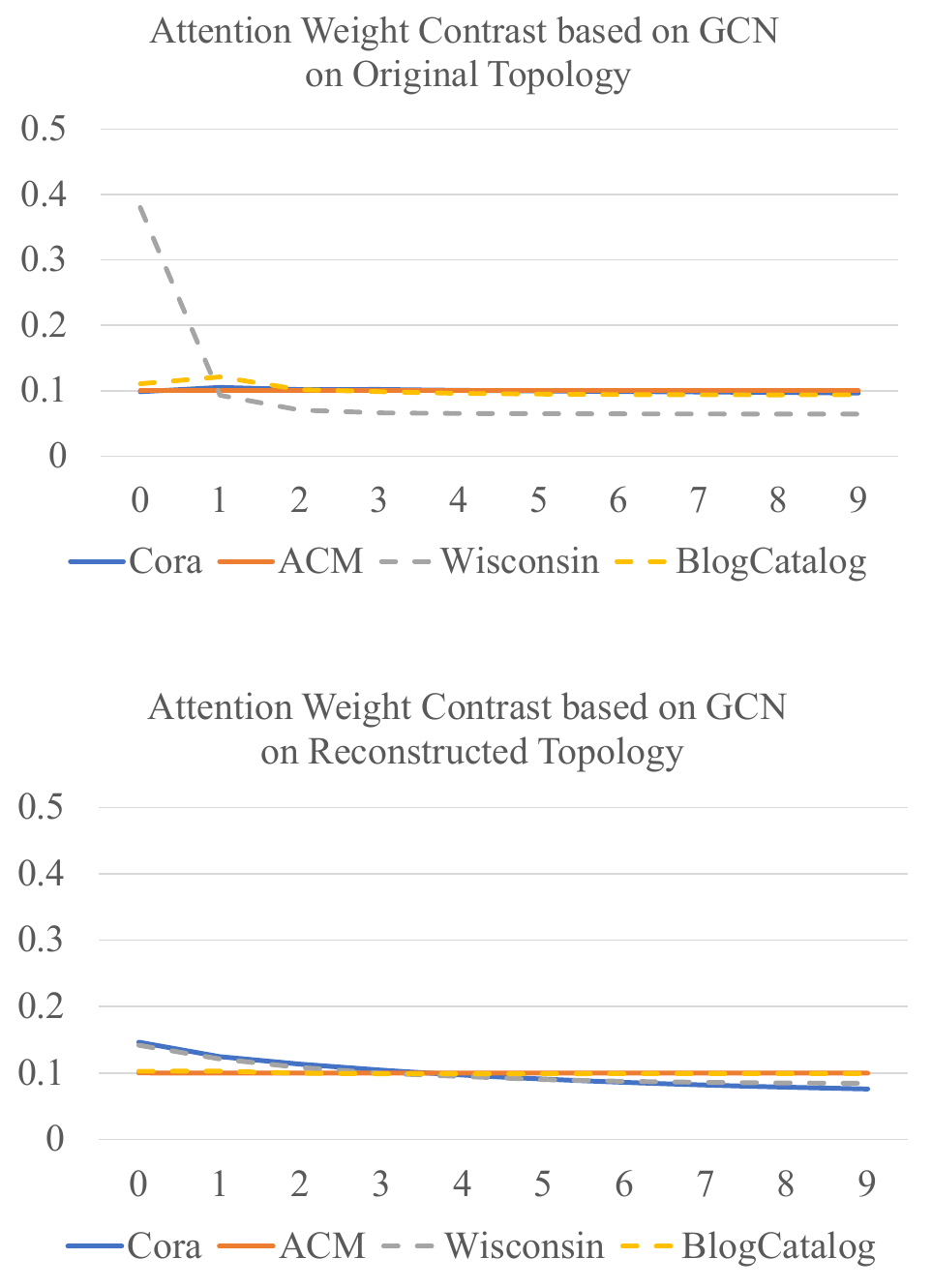}
    \caption{The distribution of attention weights across the layers of the GCN predictor for two different topologies. y-axis is attention wights, x-axis is the massage passing rounds.}
    \label{att2}
\end{figure}

The distribution of attention weights across the layers of the MLP predictor for two different topologies can be seen in Figure \ref{att1}, and that of the GCN predictor for two different topologies can be seen in Figure \ref{att2}.

\begin{figure}[!h]
    \centering
    \includegraphics[width=0.85\linewidth]{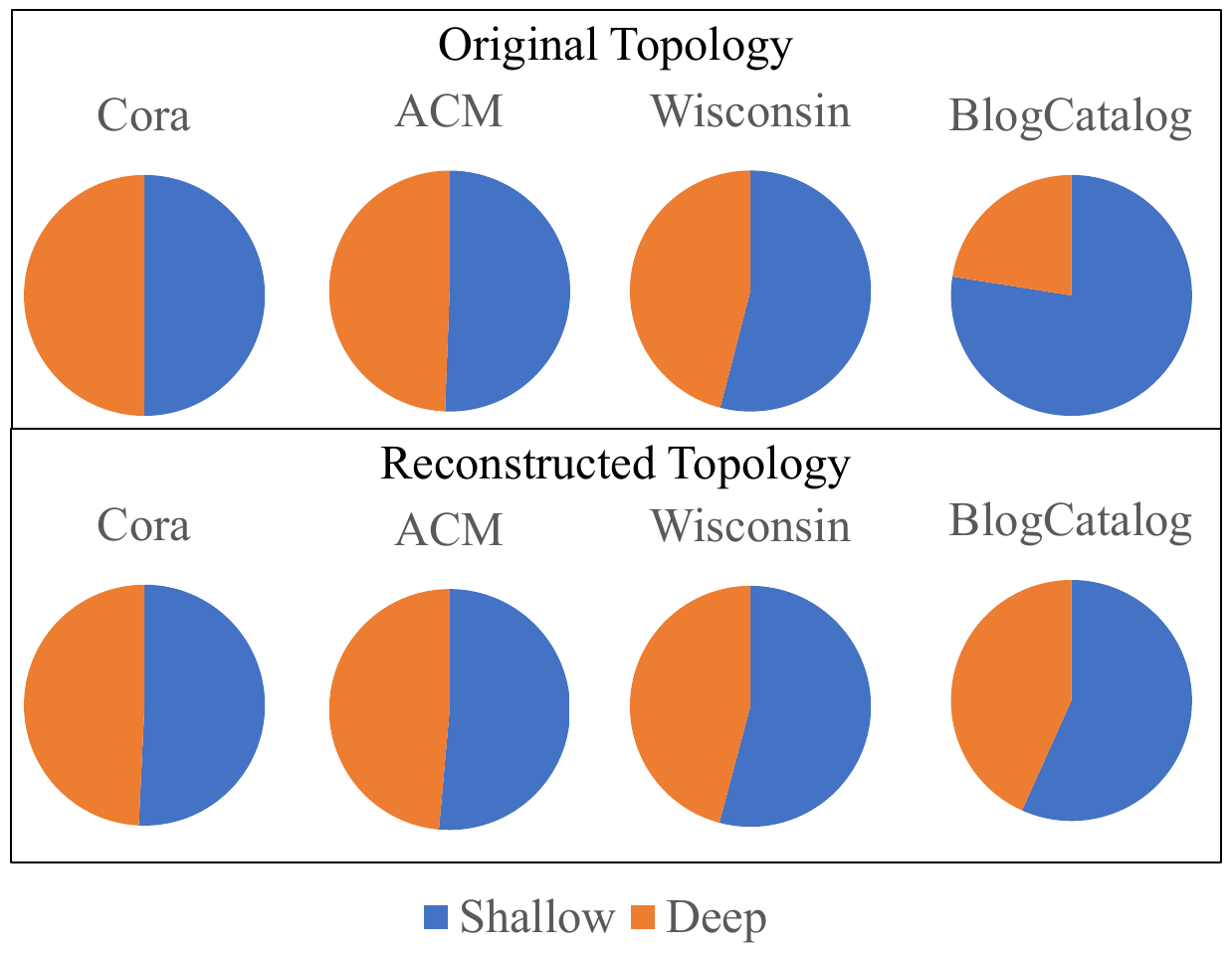}
    \caption{The proportion of attention weights in shallow and deep layers of the MLP predictor for two different topologies.
    }
    \label{att3}
\end{figure}

\begin{figure}[!h]
    \centering
    \includegraphics[width=0.85\linewidth]{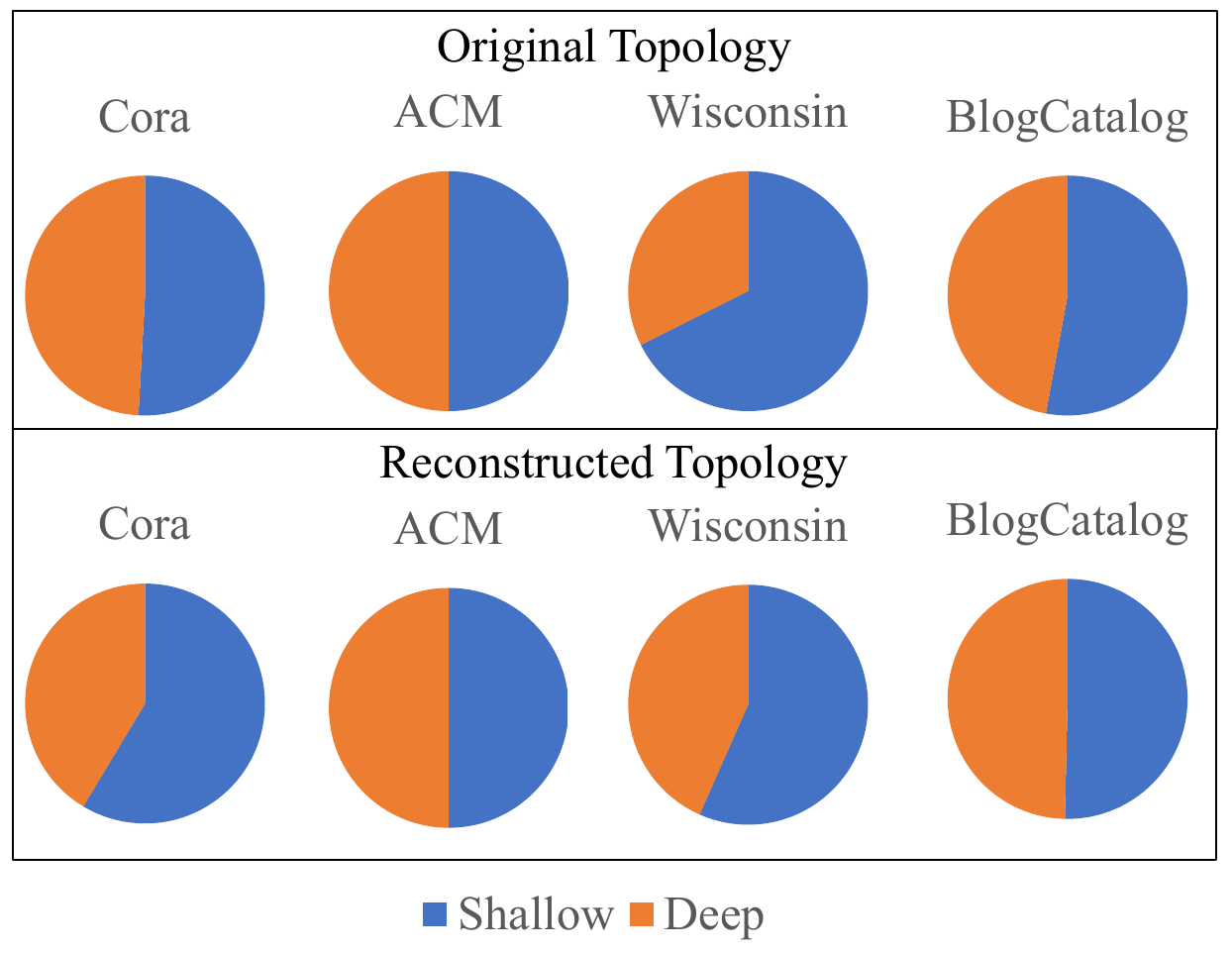}
    \caption{The proportion of attention weights in shallow and deep layers of the GCN predictor for two different topologies.
    }
    \label{att4}
\end{figure}

More intuitively, the proportions of shallow (0-4) and deep (5-9) layer embeddings in final fused representation of the MLP predictor for two different topologies are compared, as shown in Figure \ref{att3}, and that of the GCN predictor for two different topologies are compared in Figure \ref{att4}.

% \subsection{Experiment Analysis}
The results clearly demonstrate that heterophilic graphs exhibit significantly stronger attention concentration in the shallow layer compared to homophilic graphs  (e.g., the attention weight distribution of BlogCatalog dataset obtained by MLP using the original topology in Figure \ref{att1}).  Across all used heterophilic datasets, the attention weights assigned to the initial propagation layers dominate the final fused representation, confirming that attention-based convex fusion tends to collapse toward shallow semantics when deep embeddings are corrupted by structural noise.

When switching from the original topology to the reconstructed topology, we observe a consistent alleviation of shallow-layer collapse in heterophilic graphs (e.g., the attention weight distribution of BlogCatalog dataset obtained by MLP using the reconstructed topology in Figure \ref{att1}). The reconstructed topology reduces cross-class noise and preserves the semantic validity of deeper embeddings, allowing attention to be more evenly distributed across layers. This verifies that the reconstructed topology plays the intended role of mitigating misaggregation in heterophilic settings.

Interestingly, on homophilic graphs, the reconstructed topology tends to make attention more biased toward shallow layers (e.g., the attention weight distribution of Cora dataset obtained by GCN using the reconstructed topology in Figure \ref{att4}). This is because the additional edges increase local density and accelerate smoothing, making deeper embeddings less informative. Importantly, this is not detrimental to our model: the  DTPS strategy in LEDF-GNN automatically adjusts the node-wise fusion weights $\alpha$ and $\beta$, suppressing the reconstructed topology when it becomes overly smoothing. This demonstrates that DTPS is essential for adapting to unknown and diverse homophily conditions, preventing negative propagation while retaining the benefits of reconstructed topology where needed.

\section{The Details of Dataset}

\begin{table*}[!ht]
\centering
\begin{tabular}{ccccccc}
\hline
Dataset&Nodes&Edges&Features&Class&Homo.&Train/Valid/Test \\
\hline
Cora&2708&10556&1433&7&0.81&140/500/1000\\
CiteSeer&3327&9104&3703&6&0.74&120/500/1000\\
PubMed&19717&88648&500&3&0.80&60/500/1000\\
ACM&3025&26256&1870&3&0.82&60/600/1200\\
Coauthor CS&18333&163788&6805&15&0.81&300/500/1000\\
Arxiv2023&46198&78548&300&40&0.65&27718/9239/9239\\
MNIST&2000&26588&30&10&0.87&200/300/1500\\
Arxiv&169343&2315598&128&40&0.65&90941/29799/48603\\
\hline
BlogCatalog&5196&343486&8189&6&0.40&120/500/1000\\
Texas&183&325&1703&5&0.11&10/50/100\\
Wisconsin&251&515&1703&5&0.21&10/50/100\\
Cornell&183&298&1703&5&0.30&10/50/100\\
Chameleon&2277&36101&2325&5&0.23&114/500/1000\\
Squirrel&5201&217073&2089&5&0.22&260/500/1000\\
\hline
\end{tabular}
\caption{The details of fourteen datasets used in this paper. 'Homo.' refers to the homophily ratio, which is the proportion of homophily edges relative to the total number of edges.}
\label{dataset}
\end{table*}

In total, fourteen datasets are used in this work, with each being fixedly divided. The details are provided in Table \ref{dataset}.

\section{Parameter Setup}
Table \ref{paremeter} introduces the parameter settings for the semi-supervised node classification experiments, including the  $k$ parameter of $Top_k$ in LSC, as well as the propagation depths $Q1$ and $Q2$ used in the LEDF operator. Note that for both the  ablation and validation experiments, the parameter settings are identical to those used in the semi-supervised node classification experiment.

\begin{table*}[!h]
\centering
\begin{tabular}{c|c|cc|cc|cc|cc|cc}
\hline
\multirow{2}{*}{Dataset}&\multirow{2}{*}{$k$}&\multicolumn{2}{|c|}{MLP}&\multicolumn{2}{c|}{GCN}&\multicolumn{2}{c|}{GAT}&\multicolumn{2}{c|}{GIN}&\multicolumn{2}{c}{APPNP}\\
&& Q1 & Q2 & Q1 & Q2  & Q1 & Q2 & Q1 & Q2 & Q1 & Q2  \\ 
\hline
Cora&2&10&3&15&11&3&5&3&10&7&10\\
CiteSeer&3&10&9&6&20&8&1&3&10&2&7\\
PubMed&2&2&7&7&10&5&10&6&6&3&9\\
ACM&7&1&9&7&18&7&9&7&7&1&10\\
Coauthor CS&10&3&10&10&10&1&3&2&4&10&1\\
Arxiv2023&10&7&8&3&9&5&2&10&4&16&14\\
MNIST&10&7&10&6&9&9&7&4&9&5&5\\
Arxiv&10&9&6&4&6&10&4&6&6&5&8\\

\hline

BlogCatalog&20&3&8&16&14&1&10&5&9&10&9\\
Texas&20&3&8&8&16&8&20&6&8&7&4\\
Wisconsin&20&2&10&11&15&4&5&4&20&1&4\\
Cornell&20&9&7&18&2&9&8&16&10&1&17\\
chameleon&40&7&3&10&16&19&2&3&7&20&13\\
Squirrel&40&2&10&10&18&7&9&5&9&10&3\\
\hline
\end{tabular}
\caption{The parameter settings of LEDF-GNN are unified across all backbones and used consistently in the semi-supervised node classification, ablation, and validation experiments.
}
\label{paremeter}
\end{table*}

\begin{figure*}[!h]
    \centering
    \includegraphics[width=0.95\linewidth]{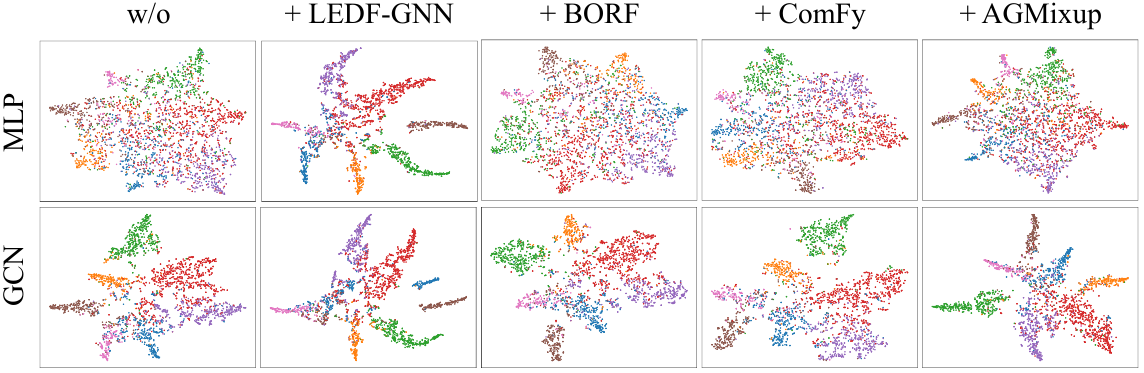}
    \caption{The visualization comparison of the backbones (MLP and GCN) and the backbones with baselines (LEDF-GNN, BORF, ComFy and AGMixup) on the \textbf{Cora} dataset.}
    \label{v1}
\end{figure*}

\section{Visualization on Semi-supervised Node Classification }

\begin{figure*}[!h]
    \centering
    \includegraphics[width=0.95\linewidth]{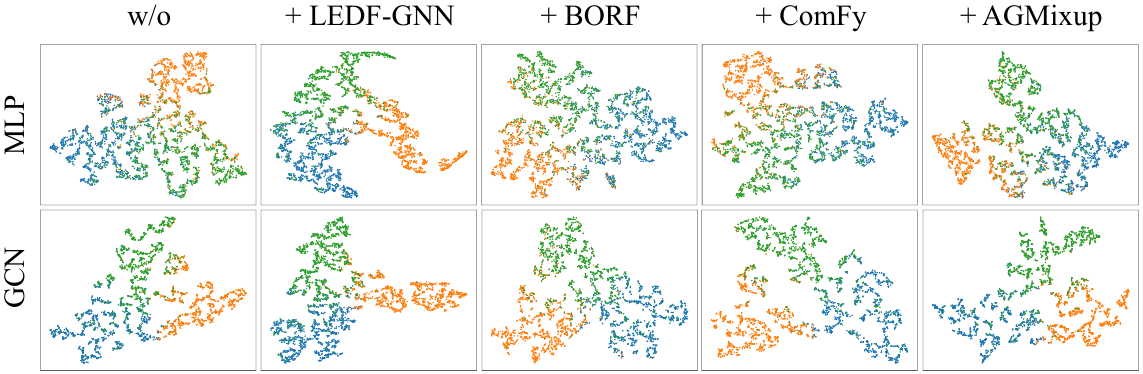}
    \caption{The visualization comparison of the backbones (MLP and GCN) and the backbones with baselines (LEDF-GNN, BORF, ComFy and AGMixup) on the \textbf{ACM} dataset.}
    \label{v2}
\end{figure*}

\begin{figure*}[!h]
    \centering
    \includegraphics[width=0.8\linewidth]{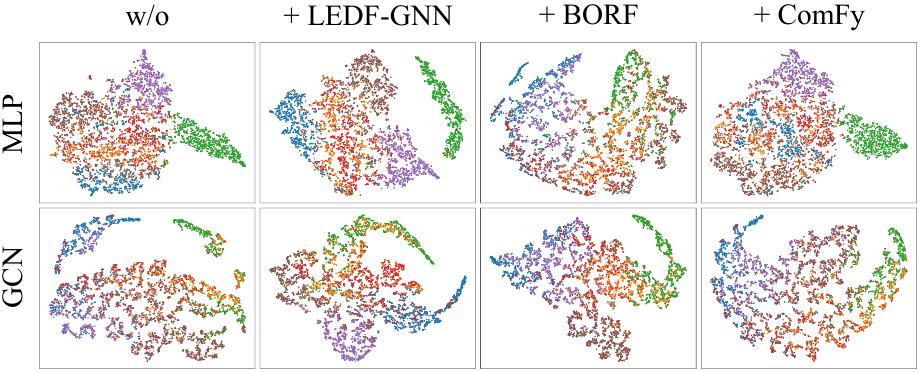}
    \caption{The visualization comparison of the backbones (MLP and GCN) and the backbones with baselines (LEDF-GNN, BORF and ComFy) on the \textbf{BlogCatalog} dataset.}
    \label{v3}
\end{figure*}

\begin{figure*}[!h]
    \centering
    \includegraphics[width=0.95\linewidth]{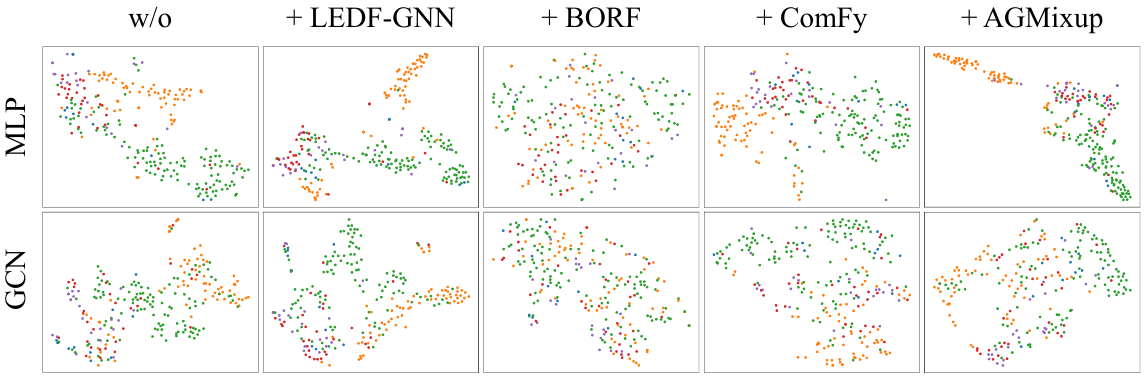}
    \caption{The visualization comparison of the backbones (MLP and GCN) and the backbones with baselines (LEDF-GNN, BORF, ComFy and AGMixup) on the \textbf{Wisconsin} dataset.}
    \label{v4}
\end{figure*}

Figure~\ref{v1} illustrates the qualitative comparison on the Cora dataset, showcasing the visual differences among the backbones (MLP and GCN) themselves, backbones with baselines (LEDF-GNN, BORF, ComFy and AGMixup). Figure~\ref{v2} illustrates the qualitative comparison on the ACM dataset, showcasing the visual differences among the backbones (MLP and GCN) themselves, backbones with baselines (LEDF-GNN, BORF, ComFy and AGMixup). Figure~\ref{v3} illustrates the qualitative comparison on the BlogCatalog dataset, showcasing the visual differences among the backbones (MLP and GCN) themselves, backbones with baselines (LEDF-GNN, BORF and ComFy). In particular, due to the complexity of AGMixup, we cannot get the result of AGMixup classification on the BlogCatalog dataset. Figure~\ref{v4} illustrates the qualitative comparison on the Wisconsin dataset, showcasing the visual differences among the backbones (MLP and GCN) themselves, backbones with baselines (LEDF-GNN, BORF, ComFy and AGMixup).

\section{LSC $vs.$ Cosine Similarity}
Table \ref{tab.vs} shows that the more intuitive LSC can obtain better homophily than cosine similarity. Bit operations based LSC needs lower memory overhead than floating-point cosine similarity within comparable time.  
\begin{table*}[!h]
\centering
\begin{tabular}{c|cc|cc|cc|cc}
\hline
\multirow{2}{*}{Dataset}&\multicolumn{2}{c|}{Empty(\%)}&\multicolumn{2}{c|}{Rewire Origin(\%)}&\multicolumn{2}{c|}{Runtime(second)}&\multicolumn{2}{c}{Memory(MB)}\\
&Cosine&Logic&Cosine&Logic&Cosine&Logic&Cosine&Logic\\
\hline
Cora&61.8&\textbf{63.2}&71.1&\textbf{75.0}&0.1057&\textbf{0.0931}&93.93&\textbf{51.12}\\
CiteSeer&64.0&\textbf{64.7}&67.0&\textbf{69.5}&\textbf{0.1489}&0.3464&186.76&\textbf{97.52}\\
Chameleon&24.2&\textbf{30.0}&23.7&\textbf{28.4}&\textbf{0.1012}&0.1082&89.10&\textbf{48.90}\\
Wisconsin&63.9&\textbf{66.6}&60.5&\textbf{62.6}&0.09816&\textbf{0.0035}&11.88&\textbf{10.01}\\
\hline
\end{tabular}
\caption{Edge homophily ratio(\%) of topologies generated from an empty graph and topologies modified from the original graph based on cosine similarity and logical similarity.}
\label{tab.vs}
\end{table*}

\section{Complexity Analysis}
The time and space complexity of our method beyond the backbone model are $O(cm+n)$ and $O(cn)$, respectively, where $n$ is node number, $m$ is edge number and $c$ is class number. Accordingly, the runtime and memory overhead reported in Tables \ref{table3} also exhibit that our model only needs lightweight overhead beyond the backbone. 
\begin{table*}[!h]
\centering
\begin{tabular}{c|cccc|cccc}
\hline
\multirow{2}{*}{Dataset}&\multicolumn{4}{c|}{Runtime(Second)}&\multicolumn{4}{c}{Memory(MB)}\\\
&GCN&BORF&AGM.&Ours&GCN&BORF&AGM.&Ours\\
\hline
Cora&0.0051&0.0119&0.1535&0.0124&49.35&49.44&351.62&50.36\\\
CiteSeer&0.0058&0.0114&0.1298&0.0178&84.74&84.90&302.70&86.75\\
Chameleon&0.0061&0.0118&0.1446&0.0147&108.87&109.56&9574.70&111.88\\
Wisconsin&0.0057&0.0141&0.0234&0.0133&22.08&22.09&66.85&22.68\\
\hline
\end{tabular}
\setlength{\abovecaptionskip}{2pt}
\caption{The runtime (Second) and Memory (MB) of backbone GCN, baselines and LEDF-GNN with backbone.}
\label{table3}
\end{table*}

\section{Hyperparameter Analysis}
\begin{figure*}[!h]
    \centering
    \includegraphics[width=0.8\linewidth]{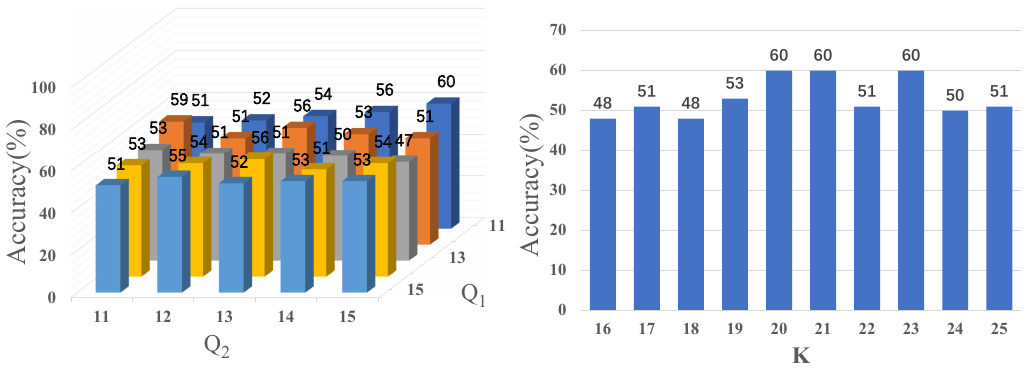}
    \caption{Sensitivity Analysis of two hyperparameters ($Q$ and $K$) on heterophilic Wisconsin dataset, one changes the other is fixed.}
    \label{sensi}
\end{figure*}

We conduct sensitivity analysis on two key hyperparameters, depth $Q$ and reconstruction hyperparameter $K$, by using the heterophilic Wisconsin dataset. Figure \ref{sensi} shows the robustness of our method across different settings.

\section{Comparison with Heterophliy-focused and Deep-GNN Baseline}
The comparison results are reported in Table \ref{tab.comparsion}. JKNet-Mean focus on layer fusion, GCNII and NDLS are deep-GNNs, H$_2$GCN is the representative heterophily-focused method. The datasets of Wisconsin and Chameleon are heterophilic.
\begin{table*}[!h]
\centering
\begin{tabular}{c|cccccc}
\hline
Model&GCN&JKNet-Mean&GCNII&NDLS&H$_2$GCN&LEDF-GCN\\
\hline
Cora&82.4&67.0&\textbf{84.3}&83.5&75.5&\underline{84.0}\\
CiteSeer&70.7&52.2&\underline{73.1}&71.5&64.1&\textbf{74.8}\\
PubMed&79.2&75.2&\underline{79.7}&79.2&77.2&\textbf{82.1}\\
ACM&82.1&76.5&\underline{82.3}&82.2&73.7&\textbf{87.5}\\
BlogCatalog&42.0&49.8&20.4&\underline{71.9}&49.0&\textbf{79.6}\\
Texas&55.0&54.0&55.0&\underline{63.0}&55.0&\textbf{66.0}\\
Wisconsin&48.0&57.0&55.0&\underline{58.0}&53.0&\textbf{60.0}\\
Cornell&\underline{49.0}&\textbf{55.0}&47.0&\textbf{55.0}&\textbf{55.0}&\textbf{55.0}\\
Chameleon&49.6&40.7&38.9&41.9&\underline{50.4}&\textbf{53.2}\\
Squirrel&\underline{33.7}&32.7&27.3&24.6&26.3&\textbf{39.0}\\
\hline
\end{tabular}
\caption{Node classification accuracy (\%) of our model and baselines, \textbf{bold} is the best, \underline{underline} is the second-best.}
\label{tab.comparsion}
\end{table*}

\end{document}